\documentclass{article}

\PassOptionsToPackage{round}{natbib}

\usepackage[dblblindworkshop, final]{neurips_2026}

\usepackage[utf8]{inputenc}
\usepackage[T1]{fontenc}
\usepackage{url}
\usepackage{booktabs}
\usepackage{amsfonts}
\usepackage{nicefrac}
\usepackage{microtype}
\usepackage{xcolor}
\usepackage{graphicx}
\usepackage{pgfplots}
\pgfplotsset{compat=1.18}
\usepgfplotslibrary{groupplots}

\usepackage{float}
\usepackage[most]{tcolorbox}

\usepackage[pagebackref=true]{hyperref}
\renewcommand*\backref[1]{\ifx#1\relax \else (Cited on page #1) \fi}
\definecolor{mygreen}{HTML}{009900}
\definecolor{deltagreen}{HTML}{308A45}
\definecolor{deltapurple}{HTML}{682075}
\definecolor[named]{ACMBlue}{cmyk}{1,0.1,0,0.1}
\definecolor[named]{ACMYellow}{cmyk}{0,0.16,1,0}
\definecolor[named]{ACMOrange}{cmyk}{0,0.42,1,0.01}
\definecolor[named]{ACMRed}{cmyk}{0,0.90,0.86,0}
\definecolor[named]{ACMLightBlue}{cmyk}{0.49,0.01,0,0}
\definecolor[named]{ACMGreen}{cmyk}{0.20,0,1,0.19}
\definecolor[named]{ACMPurple}{cmyk}{0.55,1,0,0.15}
\definecolor[named]{ACMDarkBlue}{cmyk}{1,0.58,0,0.21}
\definecolor[named]{ACMTangerine}{cmyk}{0,0.60,0.95,0.01}
\hypersetup{
  colorlinks,
  linkcolor=ACMDarkBlue,
  citecolor=ACMDarkBlue,
  urlcolor=black,
  filecolor=ACMDarkBlue,
}
\usepackage[nameinlink]{cleveref}

\definecolor{findingframe}{HTML}{1F4E79}
\definecolor{findingback}{HTML}{EDF3FA}
\newtcolorbox{finding}{
  enhanced, breakable,
  colback=findingback,
  colframe=findingframe,
  boxrule=0.2pt, leftrule=2.5pt,
  arc=2pt, outer arc=2pt,
  left=7pt, right=7pt, top=5pt, bottom=5pt,
  before skip=9pt, after skip=9pt,
  before upper={\textbf{\textcolor{findingframe}{Finding.}}\ }
}

\title{How (and How Not) to Use Data Augmentation\\ in VLA Post-Training}

\workshoptitle{RoboPAD: Post-Training Adaptation of Robot Foundation Models}

\author{%
    Bram Grooten \\
    TU Eindhoven \\
    \And
    Joaquin Vanschoren \\
    TU Eindhoven \\
}

\begin{document}

\maketitle

\begin{abstract}
Vision-language-action (VLA) models currently demonstrate strong performance in a wide range of real-world robotics tasks. However, they often still lack the generalization ability to handle large visual out-of-distribution shifts. Post-training of VLAs with reinforcement learning (RL) has been shown to benefit robustness, but significant room for improvement remains. In this work, we systematically study the effect of image augmentation on VLA post-training. We find that it is crucial to augment only the critic module during RL updates, while leaving the actor's input clean during both rollouts and updates. For $\pi_{0.5}$ and GR00T N1.5 this raises out-of-distribution success on LIBERO-Plus by $7.8$ and $10.0$ points respectively, while augmenting the actor collapses training entirely. We investigate a range of augmentation types and strengths, and provide practical recommendations for improving generalization in VLA post-training.\footnote{Code and videos are available at: \url{https://bramgrooten.nl/vla-augm/}}
\end{abstract}

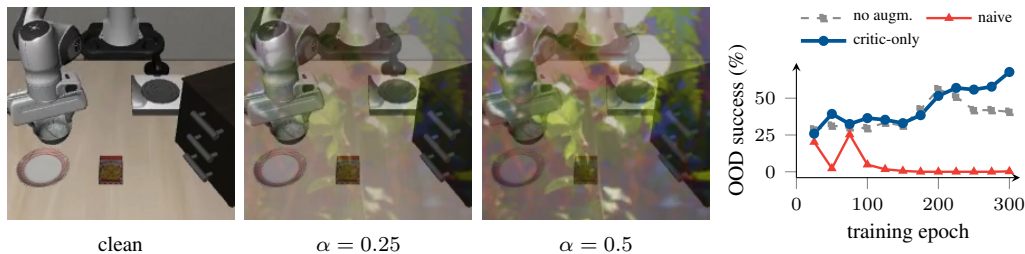
\begin{figure}[H]
\centering
\newlength{\obsw}\setlength{\obsw}{0.215\linewidth}
\begin{minipage}[b]{\obsw}\centering
  \includegraphics[width=\linewidth]{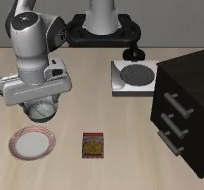}\\[1pt]
  {\footnotesize clean}
\end{minipage}\hfill
\begin{minipage}[b]{\obsw}\centering
  \includegraphics[width=\linewidth]{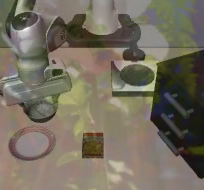}\\[1pt]
  {\footnotesize $\alpha = 0.25$}
\end{minipage}\hfill
\begin{minipage}[b]{\obsw}\centering
  \includegraphics[width=\linewidth]{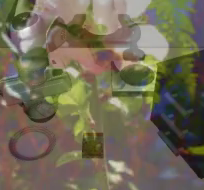}\\[1pt]
  {\footnotesize $\alpha = 0.5$}
\end{minipage}\hfill
\begin{minipage}[b]{0.325\linewidth}
  \begin{tikzpicture}
\begin{axis}[
  width=\linewidth, height=3.1cm,
  xlabel={training epoch}, ylabel={OOD success (\%)},
  xlabel style={font=\footnotesize, yshift=2pt},
  ylabel style={font=\footnotesize, yshift=-4pt},
  tick label style={font=\scriptsize},
  xmin=0, xmax=315, ymin=-4, ymax=74,
  xtick={0,100,200,300}, ytick={0,25,50,75},
  legend cell align=left, legend columns=2,
  legend style={font=\tiny, at={(0.5,1.03)}, anchor=south,
                draw=none, fill=white,
                row sep=-1pt, inner sep=2pt},
  axis lines=left, tick align=outside,
  every axis plot/.append style={mark size=1.2pt},
]
\addplot[color=black!45, mark=square*, dashed, line width=0.8pt] coordinates {(25,28.77) (50,31.34) (75,30.20) (100,29.63) (125,33.33) (150,31.34) (175,42.45) (200,56.13) (225,51.00) (250,41.88) (275,41.88) (300,40.74)};
\addlegendentry{no augm.}
\addplot[color=ACMRed, mark=triangle*, solid, line width=0.9pt] coordinates {(25,20.23) (50,2.28) (75,25.36) (100,4.84) (125,1.71) (150,0.57) (175,0.00) (200,0.00) (225,0.00) (250,0.00) (275,0.00) (300,0.28)};
\addlegendentry{naive}
\addplot[color=ACMDarkBlue, mark=*, solid, line width=1.4pt] coordinates {(25,25.93) (50,39.32) (75,32.48) (100,36.47) (125,35.33) (150,33.05) (175,38.46) (200,51.57) (225,56.98) (250,55.84) (275,57.83) (300,67.81)};
\addlegendentry{critic-only}
\end{axis}
\end{tikzpicture}
\end{minipage}
\caption{\textbf{Augmentation can improve RL post-training of VLAs.} \emph{Left:} one LIBERO observation blended with a random ImageNet image at two strengths. The VLA sees cleanly during rollouts and evaluation, but receives augmented observations for training updates. \emph{Right:} out-of-distribution success on sensor noise LIBERO-Plus environments for $\pi_{0.5}$ post-trained with PPO. Naively augmenting both actor and critic destroys the policy within 100 epochs, while giving the critic the augmented view and the actor the clean one improves OOD generalization.}
\label{fig:teaser}
\end{figure}

\section{Introduction}

Vision-language-action (VLA) models have emerged as capable generalist robot policies, yet their performance degrades sharply under visual shifts that are common in the real world: new backgrounds, lighting, distractors, and camera poses \citep{fei2025liberoplus}. A promising route to closing this gap is post-training pretrained VLAs with reinforcement learning (RL), which has been shown to improve execution and semantic generalization by \citet{liu2025rlvla}. However, the authors also indicate that RL post-training alone does \emph{not} consistently improve robustness to visual out-of-distribution (OOD) shifts, leaving a central limitation for real-world deployment unsolved.

Image augmentation is a natural, low-cost solution with a long history of improving visual robustness in deep RL, and the question of \emph{where} to inject augmentation within an actor-critic network is well studied. \citet{hansen2021svea} stabilized Q-learning by augmenting the value path while keeping the bootstrap target clean, and \citet{almuzairee2024recipe} later showed that, for standard convolutional policies, augmenting \emph{both} actor and critic inputs (with a clean target network) is more effective. Yet this line of work studies small CNN encoders with Gaussian or categorical actors, a regime far from modern vision--language--action models. In VLAs the visual encoder is a large pretrained VLM whose features risk being corrupted, and the actor is a flow-matching head through which gradients must propagate. Whether existing augmentation recipes transfer to this setting remains unknown.

In this work, we provide the first systematic study of image augmentation for VLA post-training. We ablate \emph{where} augmentation is applied (actor vs.\ critic), \emph{what} type of augmentation is used, and \emph{how strongly} it can be pushed before training collapses. Our central finding inverts prior wisdom: for VLAs post-trained with PPO, \emph{critic-only} augmentation is most effective, and augmenting the actor is detrimental to training dynamics, which is the opposite of the conclusion drawn for convolutional policies~\citep{almuzairee2024recipe}.
\Cref{fig:teaser} shows both halves of that result: what the augmented observations look like, and what applying them in the wrong place does to post-training. Based on these results, we distill practical rules of thumb for robust VLA post-training.

\section{Prior Work on Data Augmentation for VLAs}

\paragraph{Vision-language-action models and RL post-training.}
VLA models combine a pretrained vision-language backbone with an action head and are typically trained by imitation learning on large demonstration datasets~\citep{kim2024openvla, black2025pi0, black2025pi05, bjorck2025groot}.
A growing line of work further fine-tunes VLAs with online reinforcement learning~\citep{guo2025irevla, lu2026vlarl, li2026simplevlarl, zang2025rlinfvla, deng2025vlarlsurvey}, including methods designed specifically for diffusion and flow-matching action experts~\citep{ren2025dppo, zhang2025reinflow, lyu2026fpo}.
These approaches mainly improve in-distribution task success. In a controlled study, \citet{liu2025rlvla} find that RL post-training improves execution and semantic generalization, but yields little to no gain on visual out-of-distribution (OOD) shifts such as background and texture changes.
How the backbone is adapted matters too: \citet{hu2026simplerecipe} show that on-policy RL with low-rank adaptation preserves a VLA's performance in continual learning over a sequence of tasks.

\paragraph{Data augmentation in visual RL.}
Image augmentation has a long history of improving sample efficiency and generalization in visual RL~\citep{laskin2020rad, yarats2021drq, ma2025augsurvey}, and the question of \emph{how} to apply augmentation within the update has been studied in detail.
\citet{hansen2021svea} stabilize Q-learning by augmenting the value path while keeping the bootstrap target clean, \citet{raileanu2021drac} regularize policy and value toward invariance across augmentations, \citet{grooten2024madi} mask distracting image regions, and \citet{almuzairee2024recipe} conclude that augmenting \emph{both} actor and critic inputs, while keeping the target network clean, is most effective.
All of these works, however, consider small convolutional encoders with Gaussian or categorical actors, a regime far from modern VLAs with pretrained VLM backbones and flow-matching action experts.

\paragraph{Robustness and augmentation for VLAs.}
Robustness benchmarks show that VLAs are brittle under visual perturbations; on LIBERO-Plus, for instance, success rates can collapse from 95\% to below 30\% under modest lighting or background changes~\citep{liu2023libero, fei2025liberoplus}.
Several recent methods target this gap.
\citet{grover2025preserving} improve visual robustness by preserving pretrained VLM representations during fine-tuning.
On the RL side, \citet{zhang2025robustvla} add Jacobian and smoothness regularizers during post-training, \citet{xie2026strongvla} decouple robustness learning from task alignment in a two-stage curriculum, and \citet{peng2026pairvla} introduce paired auxiliary losses that enforce invariance to task-preserving visual changes and sensitivity to task-altering ones.
Other works move augmentation out of the RL update entirely: \citet{gao2026sword} apply style augmentation inside a learned world model, \citet{shi2026beyond} rely on domain-randomized simulation for sim-real co-training, and \citet{xue2025resample} augment the demonstration data itself via exploratory sampling.

\paragraph{Where should augmentation be applied?}
Despite this activity, no prior work studies \emph{where} image augmentation should be injected during RL post-training of VLAs---actor, critic, or both---even though this placement question is well understood for conventional visual RL~\citep{hansen2021svea, raileanu2021drac}.
The closest works answer adjacent questions: \citet{peng2026pairvla} achieve visual robustness in VLAs through auxiliary paired losses rather than augmentation, and \citet{almuzairee2024recipe} study augmentation placement only for small convolutional policies.
Our work fills this gap with a systematic study of where augmentation should be applied, what type works best, and how strong it can be during VLA post-training.

\begin{figure}
\centering
\newlength{\envrollw}\setlength{\envrollw}{0.178\linewidth}
\setlength{\tabcolsep}{3pt}
\begin{tabular}{@{}c@{\hspace{5pt}}ccccc@{}}
& {\small textures} & {\small lighting} & {\small layout} & {\small camera} & {\small noise} \\[3pt]
\rotatebox{90}{\footnotesize in-distribution} & \includegraphics[width=\envrollw]{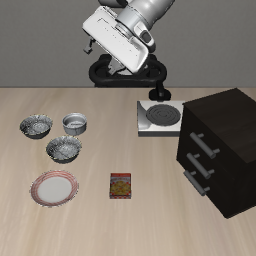} & \includegraphics[width=\envrollw]{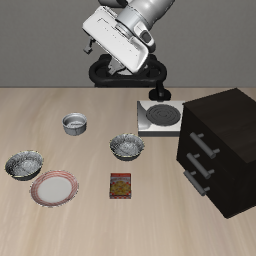} & \includegraphics[width=\envrollw]{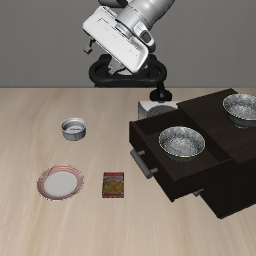} & \includegraphics[width=\envrollw]{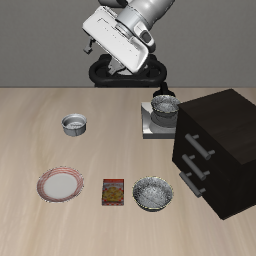} & \includegraphics[width=\envrollw]{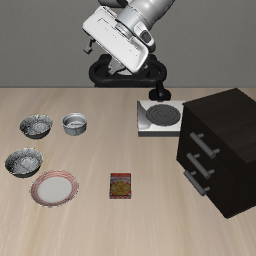} \\[5pt]
\rotatebox{90}{\footnotesize\ \ \ \ perturbed} & \includegraphics[width=\envrollw]{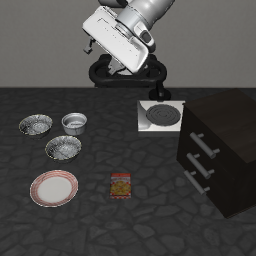} & \includegraphics[width=\envrollw]{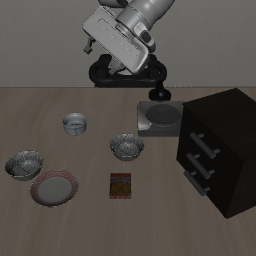} & \includegraphics[width=\envrollw]{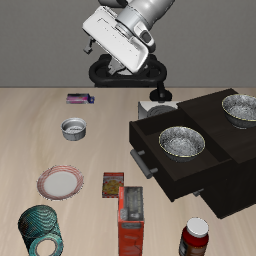} & \includegraphics[width=\envrollw]{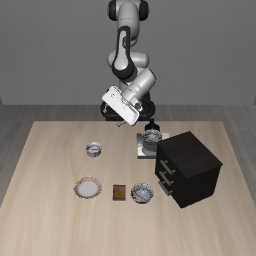} & \includegraphics[width=\envrollw]{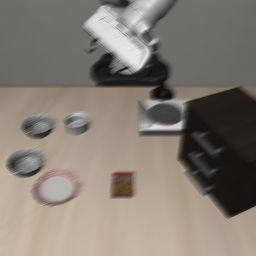} \\
\end{tabular}
\caption{\textbf{Five visual perturbation axes of LIBERO-Plus.} Each column shows the original base task on top and an out-of-distribution LIBERO-Plus perturbation below. Frames are the full main-camera observation after 20 timesteps with a random policy.}
\label{fig:envlocalroll}
\end{figure}

\section{Experimental Framework \& Augmentation Suite}
\label{sec:framework}

\paragraph{Models and post-training.}
We study two VLAs that both pair a VLM backbone with a flow-matching action head, but couple the two differently. In $\pi_{0.5}$~\citep{black2025pi05} the action expert shares one attention stack with the image and language tokens of a PaliGemma-style VLM, attending into them directly, and the robot's proprioceptive state reaches the model as discretized tokens appended to the prompt.
In GR00T N1.5~\citep{bjorck2025groot}\footnote{We use the GR00T N1.5 release; it has no separate publication, so we cite the GR00T N1 paper that introduces the architecture.} the head is instead a diffusion-transformer (DiT) that cross-attends to the features of an Eagle VLM, and proprioception enters through a separate encoder.
For both VLAs, we start from a LIBERO-Spatial supervised fine-tuning (SFT)
checkpoint and post-train it with PPO~\citep{schulman2017ppo} for up to $300$ epochs using RLinf~\citep{zang2025rlinfvla}.
Every epoch collects $512$ episodes ($8$ sequential rollout rounds over $64$ parallel environments) with each episode capped at $240$ steps.
We use a value head that takes the VLM's final features as input, so the critic and the actor share one backbone pass unless augmentation is applied to only one of them (\Cref{sec:placement}).

To run PPO on VLA models, we need the log-probability of the action taken, and a flow-matching head does not provide one. An action chunk is produced by integrating a deterministic ODE from a prior
draw $x_0 \sim \mathcal{N}(0, I)$, which gives no tractable density over actions.
We therefore use Flow-SDE~\citep{chen2025pirl}, which takes
the ODE-to-SDE conversion of Flow-GRPO~\citep{liu2025flowgrpo} and adapts it to flow-based VLAs.
Replacing the deterministic integration with an equivalent SDE that preserves the model's marginals at every timestep turns each
denoising step into a Gaussian transition whose density is available in closed
form.
The denoising chain itself becomes the stochastic policy that PPO
optimizes, and exploration comes from the injected noise.
Optimization settings, learning rates and batch sizes are listed in \Cref{sec:hyperparams}.

\paragraph{Augmentation suite.}
All augmentations act on the observation images and are applied \emph{only in the training forward pass}. Rollouts and evaluations always see clean observations, so the behavior policy and every reported success rate are measured on unaugmented inputs.
During training we adjust both camera views (base and wrist), using either \textit{overlay} and/or \textit{shift} augmentations.

\begin{figure}
\centering
\includegraphics[width=0.215\linewidth]{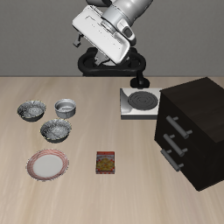}\hspace{2pt}%
\includegraphics[width=0.215\linewidth]{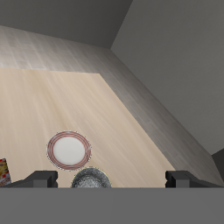}\hspace{1.5em}%
\includegraphics[width=0.215\linewidth]{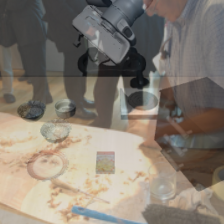}\hspace{2pt}%
\includegraphics[width=0.215\linewidth]{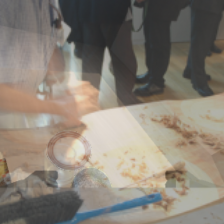}
\caption{\textbf{The observation before and after augmentation.} The base and wrist views (left pair) and the augmented versions the critic receives (right pair), blended with a random ImageNet overlay at $\alpha = 0.5$. Both views draw the same overlay image but crop and flip it independently, as in training.}
\label{fig:policyviewaug}
\end{figure}

\emph{Random overlay}~\citep{hansen2021soda}, inspired by mixup~\citep{zhang2018mixup}, convexly blends each
observation $o$ with a natural image $m$ drawn from a fixed bank,
$\tilde{o} = (1 - \alpha)\, o + \alpha\, m$, with $\alpha = 0.5$ unless stated
otherwise. The bank holds ten images from each of the $19{,}167$ classes of the ImageNet Winter21 release~\citep{deng2009imagenet}, and each draw is additionally randomly resized, cropped and flipped, so the effective overlay distribution is far larger than the image count.
A fresh overlay is drawn per sample and shared across the two camera views, so one blend roughly corresponds to a single scene change observed from both viewpoints. \Cref{fig:policyviewaug} shows both camera views of one observation before and after augmenting at the default strength. In \Cref{fig:overlay} of \Cref{sec:augmentations} we sweep the full range of $\alpha$.

\emph{Random shift}~\citep{yarats2021drq} pads the observation by $10$
pixels and takes a random crop back to the original resolution ($224{\times}224$), independently per camera as two physically mounted cameras may move separately. The pad is sized relative to the observation rather than copied outright: $10$ of $224$ pixels is a $4.5\%$ shift, matched to the $4$-pixel pad DrQ~\citep{yarats2021drq} applies to its $84{\times}84$ inputs ($4.8\%$), so the perturbation ratio stays comparable at our higher resolution. We use replicate padding, repeating the edge pixels. We also consider the composition of both augmentation types, see \Cref{sec:type}.

\paragraph{Benchmarks and metrics.}
We train on LIBERO-Spatial~\citep{liu2023libero}, a suite of ten tabletop
manipulation tasks that share objects but differ in spatial arrangement and target location.
We report in-distribution (IND) success rate on the training suite itself: each of the ten tasks is run with $50$ seeds, giving $500$ evaluation episodes per checkpoint.
To examine generalization capability, we further measure out-of-distribution (OOD) performance on LIBERO-Plus~\citep{fei2025liberoplus}, which perturbs the LIBERO tasks along seven annotated axes. We report the five \emph{visual} axes---background
textures, light conditions, objects layout, camera viewpoints and sensor
noise---comprising $1662$ of the $2402$ tasks.
Each of the $1662$ tasks is evaluated once, and we report the unweighted mean over the five axis success rates.
See \Cref{fig:envlocalroll} for examples of tasks from each LIBERO-Plus category used in this work.

\paragraph{Evaluation noise floor.}
Post-training multi-billion-parameter VLAs with PPO is costly, so we choose to run various configurations instead of seed repetitions. To investigate how much variance can exist within evaluation of the same checkpoint, we sweep one fixed model through the full evaluation suite twice and measure the absolute difference in average OOD success. This gives $0.06$ points for $\pi_{0.5}$ and $1.77$ points for GR00T N1.5. We treat a gap below this floor
as indistinguishable from evaluation noise; gaps we report are well above it.

\begin{table}
\centering
\caption{\textbf{Where augmentation is applied decides whether RL post-training survives.} We post-train VLA models on LIBERO-Spatial with PPO for 300 epochs, using random overlay augmentation at $\alpha = 0.5$. All values
are success rate percentages. \emph{Passes} counts forward passes through the VLM backbone per RL training step, and the \emph{\# tasks} row gives the suite sizes. $^\dagger$Collapsed runs were
stopped early and are reported at epoch 100.}
\label{tab:placement}
\small
\setlength{\tabcolsep}{4.5pt}
\begin{tabular}{lcccccccc}
\toprule
& & & \multicolumn{6}{c}{OOD success (LIBERO-Plus)} \\
\cmidrule(lr){4-9}
Augmented modules & Passes & IND & Tex. & Light & Layout & Cam. & Noise & Avg \\
\midrule
\emph{\# tasks} & & 10 & 258 & 292 & 385 & 376 & 351 & 1662 \\
\midrule
\multicolumn{9}{l}{\emph{$\pi_{0.5}$}~\citep{black2025pi05}} \\
SFT checkpoint (no RL)             & -- & 78.8 & 89.2 & 83.6 & 80.8 & 17.0 & 12.8 & 56.7 \\
no augm. (baseline)                & 1 & 93.2 & 95.0 & 95.9 & 92.2 & 45.7 & 40.7 & 73.9 \\
actor + critic                     & 1 & \phantom{0}0.8 & \phantom{0}0.0 & \phantom{0}0.0 & \phantom{0}0.0 & \phantom{0}0.0 & \phantom{0}0.3 & \phantom{0}0.1 \\
actor-only$^\dagger$               & 2 & 11.0 & \phantom{0}6.2 & \phantom{0}3.8 & \phantom{0}6.5 & \phantom{0}1.6 & \phantom{0}3.7 & \phantom{0}4.3 \\
critic-only                        & 2 & \textbf{98.0} & \textbf{95.8} & \textbf{96.9} & \textbf{93.0} & \textbf{55.1} & \textbf{67.8} & \textbf{81.7} \\
\midrule
\multicolumn{9}{l}{\emph{GR00T N1.5}~\citep{bjorck2025groot}} \\
SFT checkpoint (no RL)             & -- & 42.0 & 41.2 & 31.5 & 24.2 & 14.6 & 23.4 & 27.0 \\
no augm. (baseline)                & 1 & 95.4 & \textbf{90.4} & 87.7 & \textbf{62.3} & \textbf{39.4} & 50.4 & \textbf{66.0} \\
actor + critic$^\dagger$           & 1 & \phantom{0}0.2 & \phantom{0}0.0 & \phantom{0}0.0 & \phantom{0}0.3 & \phantom{0}0.3 & \phantom{0}0.0 & \phantom{0}0.1 \\
actor-only$^\dagger$               & 2 & \phantom{0}0.0 & \phantom{0}0.0 & \phantom{0}0.0 & \phantom{0}0.0 & \phantom{0}0.0 & \phantom{0}0.0 & \phantom{0}0.0 \\
critic-only                        & 2 & \textbf{97.4} & 87.3 & \textbf{88.0} & 61.6 & 34.0 & \textbf{59.3} & \textbf{66.0} \\
\bottomrule
\end{tabular}
\end{table}

\section{Where and When Does Image Augmentation Help VLAs?}
\label{sec:experiments}

We ask three questions in turn: \emph{where} augmentation should be applied in the
actor--critic update (\Cref{sec:placement}), \emph{what type} of visual augmentation works best
(\Cref{sec:type}), and \emph{how strong} it can be before post-training breaks down
(\Cref{sec:strength}).

\subsection{Which \textit{modules} should be augmented?}
\label{sec:placement}

We first investigate \emph{where} to augment, as this has been shown to be an important design question. The visual-RL literature seems to disagree on the optimal choice: SVEA augments only the critic~\citep{hansen2021svea},
DrAC regularizes both actor and critic~\citep{raileanu2021drac}, and SADA reports that
augmenting the actor helps~\citep{almuzairee2024recipe}. None of those conclusions were established for vision--language--action models.

For our experimental setup, the answer is clear: whenever the \emph{actor} receives augmented observations, training collapses; see \Cref{tab:placement}. On $\pi_{0.5}$,
augmenting both the actor and critic modules drives success down towards zero, and augmenting the actor alone was still falling when we stopped it after 100 epochs. On GR00T N1.5, both configurations which provide augmented observations to the actor also fail to train. We think this occurs because an augmented actor produces too many bad actions, leading PPO to clip an overly large fraction of the updates, as shown in \Cref{sec:training_analysis}.

Augmenting only the critic does leave post-training intact.
On $\pi_{0.5}$ it also improves on the unaugmented baseline, raising average OOD success from $73.9\%$ to $81.7\%$ ($+7.8$ points) while also
improving in-distribution success from $93.2\%$ to $98.0\%$. The per-axis columns show where it comes from: it concentrates on camera viewpoints ($45.7 \to 55.1$) and sensor noise ($40.7 \to 67.8$), while textures and lighting are already above $94\%$ without augmentation and cannot move much.
On GR00T N1.5 the same overlay leaves average OOD success unchanged at $66.0\%$. However, \Cref{sec:type} shows that GR00T N1.5 does benefit from critic-only augmentation, but from a different type (\textit{random shift} rather than \textit{overlay}).

In contrast to the findings for relatively small convolutional neural networks trained with SADA \citep{almuzairee2024recipe}, where augmenting the actor helps, we show that this is detrimental for the training dynamics of VLAs.
One downside of augmenting only the critic (or actor) module is that it requires two forward passes through the VLM backbone during post-training: a clean pass and an augmented pass. However, if memory permits, this can be concatenated along the batch dimension.

\begin{finding}
For VLAs post-trained with PPO, augmentation should be applied to the \textbf{critic only}.
\end{finding}

\begin{table}
\centering
\caption{\textbf{The best augmentation type depends on the model.} Training on LIBERO-Spatial with PPO for 300 epochs, using critic-only augmentation. All values are success rate percentages.}
\label{tab:type}
\small
\setlength{\tabcolsep}{4.5pt}
\begin{tabular}{lccccccc}
\toprule
& & \multicolumn{6}{c}{OOD success (LIBERO-Plus)} \\
\cmidrule(lr){3-8}
& IND & Tex. & Light & Layout & Cam. & Noise & Avg \\
\midrule
\multicolumn{8}{l}{\emph{$\pi_{0.5}$}~\citep{black2025pi05}} \\
no augmentation            & 93.2 & 95.0 & 95.9 & 92.2 & 45.7 & 40.7 & 73.9 \\
overlay ($\alpha = 0.5$)   & \textbf{98.0} & 95.8 & \textbf{96.9} & \textbf{93.0} & 55.1 & \textbf{67.8} & \textbf{81.7} \\
shift ($10$ px)            & 95.8 & 95.0 & 95.2 & 92.2 & \textbf{57.7} & 54.4 & 78.9 \\
overlay + shift            & 93.6 & \textbf{96.5} & 93.8 & 91.4 & 48.4 & 49.3 & 75.9 \\
\midrule
\multicolumn{8}{l}{\emph{GR00T N1.5}~\citep{bjorck2025groot}} \\
no augmentation            & 95.4 & 90.4 & 87.7 & 62.3 & 39.4 & 50.4 & 66.0 \\
overlay ($\alpha = 0.5$)   & \textbf{97.4} & 87.3 & 88.0 & 61.6 & 34.0 & \textbf{59.3} & 66.0 \\
shift ($10$ px)            & 96.4 & \textbf{93.1} & \textbf{92.5} & \textbf{77.1} & \textbf{58.8} & 58.4 & \textbf{76.0} \\
overlay + shift            & 95.4 & 87.7 & 91.4 & 67.5 & 42.3 & 55.8 & 69.0 \\
\bottomrule
\end{tabular}
\vspace{-1em}
\end{table}

\subsection{What \textit{type} of augmentation works well?}
\label{sec:type}

Holding the placement at critic-only, we compare the two augmentation types of
\Cref{sec:framework} (\textit{random overlay} and \textit{random shift}) and their combination. In \Cref{tab:type} the results show that at least one type helps on each model, but neither is uniformly better.
On $\pi_{0.5}$, overlay is the stronger of the two, raising average OOD success from $73.9\%$ to $81.7\%$ against $78.9\%$ for shift. On GR00T N1.5 the ranking reverses, with random shift reaching a $76.0\%$ success rate. The two models therefore agree that augmenting the critic helps but disagree about what to augment with.

The per-axis columns offer some insight: each augmentation naturally transfers best to the axis it most resembles.
Overlay perturbs appearance, and its largest gain is on sensor noise ($+27.1$ points on $\pi_{0.5}$, $+8.8$ on GR00T N1.5). Shift perturbs framing, and its largest gains are on the geometric axes, such as camera viewpoints ($+12.0$ and $+19.4$) and objects layout ($+14.8$ on GR00T N1.5).

The ranking between preferred augmentation types flips across models mostly because the headroom differs: $\pi_{0.5}$ is weakest on sensor noise ($40.7\%$) and GR00T N1.5 on camera viewpoints ($39.4\%$), each the specialty of the augmentation that wins for it.
Rather than fixing the best augmentation type, we recommend picking the type which perturbs the same property as the model's weakest category.
Composing the two augmentation types is not additive:
applying both seems to overly perturb the observation.
Too much augmentation can be counterproductive, which we also show in \Cref{sec:strength}.

\begin{finding}
Both overlay and shift can improve generalization, the best option is model-dependent.
\end{finding}

\begin{table}[b]
\vspace{-0.8em}
\centering
\caption{\textbf{Overlay strength has a single peak, and every setting beats no augmentation.} Training \emph{$\pi_{0.5}$}~\citep{black2025pi05} on LIBERO-Spatial with PPO for 300 epochs, using critic-only augmentation with random overlay. All values are success rate percentages.}
\label{tab:strength}
\small
\setlength{\tabcolsep}{4.5pt}
\begin{tabular}{lccccccc}
\toprule
& & \multicolumn{6}{c}{OOD success (LIBERO-Plus)} \\
\cmidrule(lr){3-8}
& IND & Tex. & Light & Layout & Cam. & Noise & Avg \\
\midrule
no augmentation            & 93.2 & 95.0 & 95.9 & 92.2 & 45.7 & 40.7 & 73.9 \\
$\alpha = 0.10$            & 96.0 & 94.6 & 94.2 & 92.7 & 50.5 & 53.0 & 77.0 \\
$\alpha = 0.25$            & 93.8 & \textbf{95.8} & 90.1 & 91.4 & 52.4 & 55.6 & 77.0 \\
$\alpha = 0.50$            & \textbf{98.0} & \textbf{95.8} & \textbf{96.9} & \textbf{93.0} & 55.1 & \textbf{67.8} & \textbf{81.7} \\
$\alpha = 0.75$            & 94.8 & 95.4 & 95.9 & 91.4 & \textbf{59.8} & 60.4 & 80.6 \\
$\alpha = 0.90$            & 93.2 & 94.6 & 93.5 & 91.9 & 51.3 & 51.6 & 76.6 \\
\bottomrule
\end{tabular}
\end{table}

\subsection{How \textit{strong} can we augment observations?}
\label{sec:strength}

We analyze the overlay strength $\alpha$ with critic-only augmentation. \Cref{tab:strength} shows that $\alpha = 0.5$ provides the best generalization performance, but other options are not far behind. Surprisingly, even at $\alpha = 0.90$, which creates extremely noisy visuals,
post-training with such augmentations benefits OOD robustness. We believe that the critic might be able to pick up on tiny pixel value deviations through the overlay of many different images over a relatively constant observation.
For $\pi_{0.5}$, average OOD success rises from $73.9\%$ to a maximum of $81.7\%$ at
$\alpha = 0.5$ and falls away on both sides, to $77.0\%$ at $\alpha = 0.10$ and
$76.6\%$ at $\alpha = 0.90$.

\begin{finding}
Moderate augmentation strength is best ($\alpha \approx 0.5$), but the differences are small.
\end{finding}

\section{Discussion}
\label{sec:discussion}

As is common in VLA post-training with RL \citep{chen2025pirl}, we freeze the vision--language backbone and train only the action expert plus a new value head on top of the VLM.
That constrains the impact that critic-only augmentation can have.
With a frozen backbone the critic's gradient cannot reshape the visual features to directly increase robustness in the actor, as done in classic visual RL \citep{hansen2021svea, grooten2024madi}.
However, critic-only augmentation does reach the actor through an improved advantage estimate $\tilde{A}_t$.
Writing $\tilde{V}$ for a critic trained on augmented observations, the temporal-difference residual and the generalized advantage estimate~\citep{schulman2016gae} the actor is trained on are
\begin{equation*}
\tilde{\delta}_t = r_t + \gamma \tilde{V}(s_{t+1}) - \tilde{V}(s_t),
\qquad
\tilde{A}_t = \sum_{l \geq 0} (\gamma \lambda)^l\, \tilde{\delta}_{t+l},
\end{equation*}
and the policy gradient is $\mathbb{E}[\nabla_\theta \log \pi_\theta(a_t \mid s_t)\, \tilde{A}_t]$.
The actor never sees an augmented pixel, and no augmented gradient reaches the VLM encoder; its update is shaped entirely by $\tilde{V}$.
A critic that scores a perturbed observation the way it scores the clean one yields more robust advantage estimates, and those estimates are the only thing the actor learns from.
The $+7.8$ points for OOD success on $\pi_{0.5}$ are provided by this mechanism.

We believe our numbers are therefore a lower bound on what critic-only augmentation can do.
Unfreezing the backbone would open a second mechanism, in which the critic's gradient flows into visual features the actor directly reads, so that augmentation shapes the representation rather than only the advantage. Repeating the settings in our work with a trainable backbone is the experiment we would run next.

\section{Conclusion}

We investigate where, with what type, and how strongly to apply image augmentation during RL post-training of vision-language-action models. The placement question is the most important: augmenting the actor's
observations collapses post-training, while applying the same augmentation only to the critic improves out-of-distribution generalization. The OOD success rate increases by $+7.8$ points on $\pi_{0.5}$ and $+10.0$ points on GR00T N1.5 at their best augmentation type.
In-distribution success does not pay for this, and in several cases improves alongside it.

\paragraph{Rules of thumb.} For practitioners post-training a VLA with PPO aiming to increase their model's out-of-distribution robustness with data augmentation:
\begin{enumerate}
    \item \textbf{Augment the critic, not the actor.} This is the one choice
    that has to be right. Feeding augmented observations to the actor drives
    training success to zero on both VLA models, whereas every critic-only augmentation shows stable performance.
    \item \textbf{Match the augmentation to your model's weakest axis.} Overlay perturbs appearance and helps most on sensor noise; shift perturbs framing and helps most on camera viewpoints and layout. Overlay won for $\pi_{0.5}$ and shift for GR00T N1.5 because that is where each had most to gain. Combining augmentations was worse than the better one alone.
    \item \textbf{Keep the augmentation strength moderate.} Augmenting with $\alpha \approx 0.5$ was best, but the
    curve is flat enough that every value from $0.1$ to $0.9$ still beats the unaugmented baseline.
\end{enumerate}

\paragraph{Limitations.} Our study is bounded in scope: all runs are in simulation, on one benchmark, and with one RL algorithm, so whether the conclusions hold on real hardware, on other task suites, or with off-policy algorithms remains untested.
Within that scope we chose coverage over repetition, spanning two VLAs with three augmentation types and five overlay strengths applied to four different parts of the models. This coverage costs us one training seed per configuration, though each run is scored on over a thousand LIBERO-Plus perturbations.
Our conclusions do not depend on a single run: augmenting the actor collapses training on both models, and the strength sweep traces a curve across five values.
Finally, we augment images only: varying language instructions and creating noisy proprioceptive inputs is left to future work.

\paragraph{Outlook.} The practical message is encouraging:
image augmentation can readily be applied to RL post-training of VLAs, as long as it is done right (only on the critic, only during updates).
Augmentation improves out-of-distribution
success on both VLAs we tested, while maintaining in-distribution performance, making it a low-risk default. That the same critic-only placement wins across two rather
different VLA model structures, one sharing an attention stack with the VLM ($\pi_{0.5}$) and one cross-attending
to it (GR00T N1.5), suggests the effect follows from the common actor-critic structure itself rather than from any single architecture. We hope this makes RL post-training a more dependable route to visually robust robot policies.

\section*{Acknowledgements}

Funded by the European Union. Views and opinions expressed are however those of the author(s) only and do not necessarily reflect those of the European Union or the European Commission. Neither the European Union nor the European Commission can be held responsible for them. This work was supported by the European Union's Horizon Europe research and innovation programme under grant agreement No 101214398 (ELLIOT).
This publication is part of project 2026.008 of the Computing Time National Computing Facilities program, which is financed by the Dutch Research Council (NWO) under the grant \url{https://doi.org/10.61686/WBFDE48237}.

\bibliographystyle{plainnat}
{\small
\bibliography{citations}

\begin{thebibliography}{37}
\providecommand{\natexlab}[1]{#1}
\providecommand{\url}[1]{\texttt{#1}}
\expandafter\ifx\csname urlstyle\endcsname\relax
  \providecommand{\doi}[1]{doi: #1}\else
  \providecommand{\doi}{doi: \begingroup \urlstyle{rm}\Url}\fi

\bibitem[Almuzairee et~al.(2024)Almuzairee, Hansen, and
  Christensen]{almuzairee2024recipe}
Abdulaziz Almuzairee, Nicklas Hansen, and Henrik~I Christensen.
\newblock {A Recipe for Unbounded Data Augmentation in Visual Reinforcement
  Learning}.
\newblock \emph{Reinforcement Learning Conference (RLC)}, 2024.
\newblock URL: \url{https://arxiv.org/abs/2405.17416}.

\bibitem[Bjorck et~al.(2025)Bjorck, Casta\~neda, Cherniadev, Da, Ding, Fan,
  Fang, Fox, Hu, Huang, Jang, Jiang, Kautz, Kundalia, Lao, Li, Lin, Lin, Liu,
  Llontop, Magne, Mandlekar, Narayan, Nasiriany, Reed, Tan, Wang, Wang, Wang,
  Wang, Xiang, Xie, Xu, Xu, Ye, Yu, Zhang, Zhang, Zhao, Zheng, and
  Zhu]{bjorck2025groot}
Johan Bjorck, Fernando Casta\~neda, Nikita Cherniadev, Xingye Da, Runyu Ding,
  Linxi Fan, Yu~Fang, Dieter Fox, Fengyuan Hu, Spencer Huang, Joel Jang, Zhenyu
  Jiang, Jan Kautz, Kaushil Kundalia, Lawrence Lao, Zhiqi Li, Zongyu Lin, Kevin
  Lin, Guilin Liu, Edith Llontop, Loic Magne, Ajay Mandlekar, Avnish Narayan,
  Soroush Nasiriany, Scott Reed, You~Liang Tan, Guanzhi Wang, Zu~Wang, Jing
  Wang, Qi~Wang, Jiannan Xiang, Yuqi Xie, Yinzhen Xu, Zhenjia Xu, Seonghyeon
  Ye, Zhiding Yu, Ao~Zhang, Hao Zhang, Yizhou Zhao, Ruijie Zheng, and Yuke Zhu.
\newblock {GR00T N1: An Open Foundation Model for Generalist Humanoid Robots}.
\newblock \emph{arXiv preprint arXiv:2503.14734}, 2025.
\newblock URL: \url{https://arxiv.org/abs/2503.14734}.

\bibitem[Black et~al.(2025{\natexlab{a}})Black, Brown, Darpinian, Dhabalia,
  Driess, Esmail, Equi, Finn, Fusai, Galliker, Ghosh, Groom, Hausman, Ichter,
  Jakubczak, Jones, Ke, LeBlanc, Levine, Li-Bell, Mothukuri, Nair, Pertsch,
  Ren, Shi, Smith, Springenberg, Stachowicz, Tanner, Vuong, Walke, Walling,
  Wang, Yu, and Zhilinsky]{black2025pi05}
Kevin Black, Noah Brown, James Darpinian, Karan Dhabalia, Danny Driess, Adnan
  Esmail, Michael Equi, Chelsea Finn, Niccolo Fusai, Manuel~Y Galliker, Dibya
  Ghosh, Lachy Groom, Karol Hausman, Brian Ichter, Szymon Jakubczak, Tim Jones,
  Liyiming Ke, Devin LeBlanc, Sergey Levine, Adrian Li-Bell, Mohith Mothukuri,
  Suraj Nair, Karl Pertsch, Allen~Z Ren, Lucy~Xiaoyang Shi, Laura Smith,
  Jost~Tobias Springenberg, Kyle Stachowicz, James Tanner, Quan Vuong, Homer
  Walke, Anna Walling, Haohuan Wang, Lili Yu, and Ury Zhilinsky.
\newblock {$\pi_{0.5}$: A Vision-Language-Action Model with Open-World
  Generalization}.
\newblock \emph{Conference on Robot Learning (CoRL)}, 2025{\natexlab{a}}.
\newblock URL: \url{https://arxiv.org/abs/2504.16054}.

\bibitem[Black et~al.(2025{\natexlab{b}})Black, Brown, Driess, Esmail, Equi,
  Finn, Fusai, Groom, Hausman, Ichter, Jakubczak, Jones, Ke, Levine, Li-Bell,
  Mothukuri, Nair, Pertsch, Shi, Tanner, Vuong, Walling, Wang, and
  Zhilinsky]{black2025pi0}
Kevin Black, Noah Brown, Danny Driess, Adnan Esmail, Michael Equi, Chelsea
  Finn, Niccolo Fusai, Lachy Groom, Karol Hausman, Brian Ichter, Szymon
  Jakubczak, Tim Jones, Liyiming Ke, Sergey Levine, Adrian Li-Bell, Mohith
  Mothukuri, Suraj Nair, Karl Pertsch, Lucy~Xiaoyang Shi, James Tanner, Quan
  Vuong, Anna Walling, Haohuan Wang, and Ury Zhilinsky.
\newblock {$\pi_0$: A Vision-Language-Action Flow Model for General Robot
  Control}.
\newblock \emph{Robotics: Science and Systems (RSS)}, 2025{\natexlab{b}}.
\newblock URL: \url{https://arxiv.org/abs/2410.24164}.

\bibitem[Chen et~al.(2025)Chen, Liu, Zhang, Guo, Xu, Lin, Zang, Li, Wei, Zhou,
  Zhang, Yu, Fan, Huang, Wang, and Yu]{chen2025pirl}
Kang Chen, Zhihao Liu, Tonghe Zhang, Zhen Guo, Si~Xu, Hao Lin, Hongzhi Zang,
  Xiang Li, Bingwen Wei, Jiakai Zhou, Quanlu Zhang, Zhaofei Yu, Guoliang Fan,
  Tiejun Huang, Yu~Wang, and Chao Yu.
\newblock {$\pi_\texttt{RL}$: Online RL Fine-tuning for Flow-based
  Vision-Language-Action Models}.
\newblock \emph{arXiv preprint arXiv:2510.25889}, 2025.
\newblock URL: \url{https://arxiv.org/abs/2510.25889}.

\bibitem[Deng et~al.(2025)Deng, Wu, Liu, Guo, Xue, Shan, Zhang, Jia, Ling, Lu,
  and Wang]{deng2025vlarlsurvey}
Haoyuan Deng, Zhenyu Wu, Haichao Liu, Wenkai Guo, Yuquan Xue, Ziyu Shan,
  Chuanrui Zhang, Bofang Jia, Yuan Ling, Guanxing Lu, and Ziwei Wang.
\newblock {A Survey on Reinforcement Learning of Vision-Language-Action Models
  for Robotic Manipulation}.
\newblock \emph{TechRxiv preprint}, 2025.
\newblock URL:
  \url{https://www.techrxiv.org/doi/full/10.36227/techrxiv.176531955.54563920/v1}.

\bibitem[Deng et~al.(2009)Deng, Dong, Socher, Li, Li, and
  Fei-Fei]{deng2009imagenet}
Jia Deng, Wei Dong, Richard Socher, Li-Jia Li, Kai Li, and Li~Fei-Fei.
\newblock {ImageNet: A Large-Scale Hierarchical Image Database}.
\newblock \emph{IEEE Conference on Computer Vision and Pattern Recognition
  (CVPR)}, 2009.
\newblock URL: \url{https://ieeexplore.ieee.org/document/5206848}.

\bibitem[Fei et~al.(2025)Fei, Wang, Shi, Dai, Cai, Qian, Ji, He, Zhang, Fei,
  Fu, Gong, and Qiu]{fei2025liberoplus}
Senyu Fei, Siyin Wang, Junhao Shi, Zihao Dai, Jikun Cai, Pengfang Qian, Li~Ji,
  Xinzhe He, Shiduo Zhang, Zhaoye Fei, Jinlan Fu, Jingjing Gong, and Xipeng
  Qiu.
\newblock {LIBERO-Plus: In-Depth Robustness Analysis of Vision-Language-Action
  Models}.
\newblock \emph{arXiv preprint arXiv:2510.13626}, 2025.
\newblock URL: \url{https://arxiv.org/abs/2510.13626}.

\bibitem[Gao et~al.(2026)Gao, Guo, Guan, Huang, Ma, Xiao, Xiong, and
  Wen]{gao2026sword}
Jiaxuan Gao, Yongjian Guo, Zhong Guan, Wen Huang, Wanlun Ma, Xi~Xiao, Junwu
  Xiong, and Sheng Wen.
\newblock {Sword: Style-Robust World Models as Simulators via Dynamic Latent
  Bootstrapping for VLA Policy Post-Training}.
\newblock \emph{arXiv preprint arXiv:2605.07288}, 2026.
\newblock URL: \url{https://arxiv.org/abs/2605.07288}.

\bibitem[Grooten et~al.(2024)Grooten, Tomilin, Vasan, Taylor, Mahmood, Fang,
  Pechenizkiy, and Mocanu]{grooten2024madi}
Bram Grooten, Tristan Tomilin, Gautham Vasan, Matthew~E Taylor, A~Rupam
  Mahmood, Meng Fang, Mykola Pechenizkiy, and Decebal~Constantin Mocanu.
\newblock {MaDi: Learning to Mask Distractions for Generalization in Visual
  Deep Reinforcement Learning}.
\newblock \emph{Int. Conf. Autonomous Agents and Multiagent Systems (AAMAS)},
  2024.
\newblock URL: \url{https://arxiv.org/abs/2312.15339}.

\bibitem[Grover et~al.(2025)Grover, Gopalkrishnan, Ai, Christensen, Su, and
  Li]{grover2025preserving}
Shresth Grover, Akshay Gopalkrishnan, Bo~Ai, Henrik~I Christensen, Hao Su, and
  Xuanlin Li.
\newblock {Enhancing Generalization in Vision-Language-Action Models by
  Preserving Pretrained Representations}.
\newblock \emph{arXiv preprint arXiv:2509.11417}, 2025.
\newblock URL: \url{https://arxiv.org/abs/2509.11417}.

\bibitem[Guo et~al.(2025)Guo, Zhang, Chen, Ji, Wang, Hu, and
  Chen]{guo2025irevla}
Yanjiang Guo, Jianke Zhang, Xiaoyu Chen, Xiang Ji, Yen-Jen Wang, Yucheng Hu,
  and Jianyu Chen.
\newblock {Improving Vision-Language-Action Model with Online Reinforcement
  Learning}.
\newblock \emph{IEEE International Conference on Robotics and Automation
  (ICRA)}, 2025.
\newblock URL: \url{https://arxiv.org/abs/2501.16664}.

\bibitem[Hansen and Wang(2021)]{hansen2021soda}
Nicklas Hansen and Xiaolong Wang.
\newblock {Generalization in Reinforcement Learning by Soft Data Augmentation}.
\newblock In \emph{IEEE International Conference on Robotics and Automation
  (ICRA)}, pages 13611--13617. IEEE, 2021.
\newblock URL: \url{https://arxiv.org/abs/2011.13389}.

\bibitem[Hansen et~al.(2021)Hansen, Su, and Wang]{hansen2021svea}
Nicklas Hansen, Hao Su, and Xiaolong Wang.
\newblock {Stabilizing Deep Q-Learning with ConvNets and Vision Transformers
  under Data Augmentation}.
\newblock \emph{Advances in Neural Information Processing Systems (NeurIPS)},
  2021.
\newblock URL: \url{https://arxiv.org/abs/2107.00644}.

\bibitem[Hu et~al.(2026)Hu, Shim, Tang, Sung, Liu, Stone, and
  Mart\'{i}n-Mart\'{i}n]{hu2026simplerecipe}
Jiaheng Hu, Jay Shim, Chen Tang, Yoonchang Sung, Bo~Liu, Peter Stone, and
  Roberto Mart\'{i}n-Mart\'{i}n.
\newblock {Simple Recipe Works: Vision-Language-Action Models Are Natural
  Continual Learners with Reinforcement Learning}.
\newblock \emph{Reinforcement Learning Conference (RLC)}, 2026.
\newblock URL: \url{https://arxiv.org/abs/2603.11653}.

\bibitem[Kim et~al.(2024)Kim, Pertsch, Karamcheti, Xiao, Balakrishna, Nair,
  Rafailov, Foster, Lam, Sanketi, Vuong, Kollar, Burchfiel, Tedrake, Sadigh,
  Levine, Liang, and Finn]{kim2024openvla}
Moo~Jin Kim, Karl Pertsch, Siddharth Karamcheti, Ted Xiao, Ashwin Balakrishna,
  Suraj Nair, Rafael Rafailov, Ethan Foster, Grace Lam, Pannag Sanketi, Quan
  Vuong, Thomas Kollar, Benjamin Burchfiel, Russ Tedrake, Dorsa Sadigh, Sergey
  Levine, Percy Liang, and Chelsea Finn.
\newblock {OpenVLA: An Open-Source Vision-Language-Action Model}.
\newblock \emph{Conference on Robot Learning (CoRL)}, 2024.
\newblock URL: \url{https://arxiv.org/abs/2406.09246}.

\bibitem[Laskin et~al.(2020)Laskin, Lee, Stooke, Pinto, Abbeel, and
  Srinivas]{laskin2020rad}
Michael Laskin, Kimin Lee, Adam Stooke, Lerrel Pinto, Pieter Abbeel, and
  Aravind Srinivas.
\newblock {Reinforcement Learning with Augmented Data}.
\newblock \emph{Advances in Neural Information Processing Systems (NeurIPS)},
  2020.
\newblock URL: \url{https://arxiv.org/abs/2004.14990}.

\bibitem[Li et~al.(2026)Li, Zuo, Yu, Zhang, Yang, Zhang, Zhu, Zhang, Chen, Cui,
  Wang, Luo, Fan, Sun, Zeng, Pang, Zhang, Wang, Mu, Zhou, and
  Ding]{li2026simplevlarl}
Haozhan Li, Yuxin Zuo, Jiale Yu, Yuhao Zhang, Zhaohui Yang, Kaiyan Zhang,
  Xuekai Zhu, Yuchen Zhang, Tianxing Chen, Ganqu Cui, Dehui Wang, Dingxiang
  Luo, Yuchen Fan, Youbang Sun, Jia Zeng, Jiangmiao Pang, Shanghang Zhang,
  Yu~Wang, Yao Mu, Bowen Zhou, and Ning Ding.
\newblock {SimpleVLA-RL: Scaling VLA Training via Reinforcement Learning}.
\newblock \emph{International Conference on Learning Representations (ICLR)},
  2026.
\newblock URL: \url{https://arxiv.org/abs/2509.09674}.

\bibitem[Liu et~al.(2023)Liu, Zhu, Gao, Feng, Liu, Zhu, and
  Stone]{liu2023libero}
Bo~Liu, Yifeng Zhu, Chongkai Gao, Yihao Feng, Qiang Liu, Yuke Zhu, and Peter
  Stone.
\newblock {LIBERO: Benchmarking Knowledge Transfer for Lifelong Robot
  Learning}.
\newblock \emph{Advances in Neural Information Processing Systems (NeurIPS)
  Datasets and Benchmarks}, 2023.
\newblock URL: \url{https://arxiv.org/abs/2306.03310}.

\bibitem[Liu et~al.(2025{\natexlab{a}})Liu, Liu, Liang, Li, Liu, Wang, Wan,
  Zhang, and Ouyang]{liu2025flowgrpo}
Jie Liu, Gongye Liu, Jiajun Liang, Yangguang Li, Jiaheng Liu, Xintao Wang,
  Pengfei Wan, Di~Zhang, and Wanli Ouyang.
\newblock {Flow-GRPO: Training Flow Matching Models via Online RL}.
\newblock \emph{Advances in Neural Information Processing Systems (NeurIPS)},
  2025{\natexlab{a}}.
\newblock URL: \url{https://arxiv.org/abs/2505.05470}.

\bibitem[Liu et~al.(2025{\natexlab{b}})Liu, Gao, Wei, Chen, Liao, Wu, Yu, and
  Wang]{liu2025rlvla}
Jijia Liu, Feng Gao, Bingwen Wei, Xinlei Chen, Qingmin Liao, Yi~Wu, Chao Yu,
  and Yu~Wang.
\newblock {What Can RL Bring to VLA Generalization? An Empirical Study}.
\newblock \emph{Advances in Neural Information Processing Systems (NeurIPS)},
  2025{\natexlab{b}}.
\newblock URL: \url{https://arxiv.org/abs/2505.19789}.

\bibitem[Lu et~al.(2026)Lu, Guo, Zhang, Zhou, Jiang, Gao, Tang, and
  Wang]{lu2026vlarl}
Guanxing Lu, Wenkai Guo, Chubin Zhang, Yuheng Zhou, Haonan Jiang, Zifeng Gao,
  Yansong Tang, and Ziwei Wang.
\newblock {VLA-RL: Towards Masterful and General Robotic Manipulation with
  Scalable Reinforcement Learning}.
\newblock \emph{IEEE/RSJ International Conference on Intelligent Robots and
  Systems (IROS)}, 2026.
\newblock URL: \url{https://arxiv.org/abs/2505.18719}.

\bibitem[Lyu et~al.(2026)Lyu, Sun, Lin, Li, Chen, Zhao, and Zeng]{lyu2026fpo}
Mingyang Lyu, Yinqian Sun, Erliang Lin, Huangrui Li, Ruolin Chen, Feifei Zhao,
  and Yi~Zeng.
\newblock {Reinforcement Fine-Tuning of Flow-Matching Policies for
  Vision-Language-Action Models}.
\newblock \emph{IEEE International Conference on Robotics and Automation
  (ICRA)}, 2026.
\newblock URL: \url{https://arxiv.org/abs/2510.09976}.

\bibitem[Ma et~al.(2025)Ma, Wang, Yuan, Wang, Yuan, and Tao]{ma2025augsurvey}
Guozheng Ma, Zhen Wang, Zhecheng Yuan, Xueqian Wang, Bo~Yuan, and Dacheng Tao.
\newblock {A Comprehensive Survey of Data Augmentation in Visual Reinforcement
  Learning}.
\newblock \emph{International Journal of Computer Vision (IJCV)}, 2025.
\newblock URL: \url{https://arxiv.org/abs/2210.04561}.

\bibitem[Peng et~al.(2026)Peng, Fu, Zhang, Zhao, Bian, Liu, Zhang, Zhang, and
  Wang]{peng2026pairvla}
Yuanfang Peng, Jingjing Fu, Chuheng Zhang, Li~Zhao, Jiang Bian, Mingyu Liu,
  Ling Zhang, Jun Zhang, and Rui Wang.
\newblock {What to Ignore, What to React: Visually Robust RL Fine-Tuning of VLA
  Models}.
\newblock \emph{arXiv preprint arXiv:2605.13105}, 2026.
\newblock URL: \url{https://arxiv.org/abs/2605.13105}.

\bibitem[Raileanu et~al.(2021)Raileanu, Goldstein, Yarats, Kostrikov, and
  Fergus]{raileanu2021drac}
Roberta Raileanu, Max Goldstein, Denis Yarats, Ilya Kostrikov, and Rob Fergus.
\newblock {Automatic Data Augmentation for Generalization in Reinforcement
  Learning}.
\newblock \emph{Advances in Neural Information Processing Systems (NeurIPS)},
  2021.
\newblock URL: \url{https://arxiv.org/abs/2006.12862}.

\bibitem[Ren et~al.(2025)Ren, Lidard, Ankile, Simeonov, Agrawal, Majumdar,
  Burchfiel, Dai, and Simchowitz]{ren2025dppo}
Allen~Z Ren, Justin Lidard, Lars~L Ankile, Anthony Simeonov, Pulkit Agrawal,
  Anirudha Majumdar, Benjamin Burchfiel, Hongkai Dai, and Max Simchowitz.
\newblock {Diffusion Policy Policy Optimization}.
\newblock \emph{International Conference on Learning Representations (ICLR)},
  2025.
\newblock URL: \url{https://arxiv.org/abs/2409.00588}.

\bibitem[Schulman et~al.(2016)Schulman, Moritz, Levine, Jordan, and
  Abbeel]{schulman2016gae}
John Schulman, Philipp Moritz, Sergey Levine, Michael Jordan, and Pieter
  Abbeel.
\newblock {High-Dimensional Continuous Control Using Generalized Advantage
  Estimation}.
\newblock \emph{International Conference on Learning Representations (ICLR)},
  2016.
\newblock URL: \url{https://arxiv.org/abs/1506.02438}.

\bibitem[Schulman et~al.(2017)Schulman, Wolski, Dhariwal, Radford, and
  Klimov]{schulman2017ppo}
John Schulman, Filip Wolski, Prafulla Dhariwal, Alec Radford, and Oleg Klimov.
\newblock {Proximal Policy Optimization Algorithms}.
\newblock \emph{arXiv preprint arXiv:1707.06347}, 2017.
\newblock URL: \url{https://arxiv.org/abs/1707.06347}.

\bibitem[Shi et~al.(2026)Shi, Chen, Gao, Chen, Chen, Zhang, Zang, Zhou, Zhang,
  Yu, and Wang]{shi2026beyond}
Liangzhi Shi, Shuaihang Chen, Feng Gao, Yinuo Chen, Kang Chen, Tonghe Zhang,
  Hongzhi Zang, Jiakai Zhou, Weinan Zhang, Chao Yu, and Yu~Wang.
\newblock {Beyond Imitation: Reinforcement Learning-Based Sim-Real Co-Training
  for VLA Models}.
\newblock \emph{arXiv preprint arXiv:2602.12628}, 2026.
\newblock URL: \url{https://arxiv.org/abs/2602.12628}.

\bibitem[Xie et~al.(2026)Xie, Yan, Zhao, Wang, and Jin]{xie2026strongvla}
Yuhan Xie, Yuping Yan, Yunqi Zhao, Handing Wang, and Yaochu Jin.
\newblock {STRONG-VLA: Decoupled Robustness Learning for Vision-Language-Action
  Models under Multimodal Perturbations}.
\newblock \emph{arXiv preprint arXiv:2604.10055}, 2026.
\newblock URL: \url{https://arxiv.org/abs/2604.10055}.

\bibitem[Xue et~al.(2025)Xue, Lu, Wu, Zhang, Jia, Gu, and
  Wang]{xue2025resample}
Yuquan Xue, Guanxing Lu, Zhenyu Wu, Chuanrui Zhang, Bofang Jia, Zhengyi Gu, and
  Ziwei Wang.
\newblock {RESample: A Robust Data Augmentation Framework via Exploratory
  Sampling for Robotic Manipulation}.
\newblock \emph{arXiv preprint arXiv:2510.17640}, 2025.
\newblock URL: \url{https://arxiv.org/abs/2510.17640}.

\bibitem[Yarats et~al.(2021)Yarats, Kostrikov, and Fergus]{yarats2021drq}
Denis Yarats, Ilya Kostrikov, and Rob Fergus.
\newblock {Image Augmentation Is All You Need: Regularizing Deep Reinforcement
  Learning from Pixels}.
\newblock \emph{International Conference on Learning Representations (ICLR)},
  2021.
\newblock URL: \url{https://arxiv.org/abs/2004.13649}.

\bibitem[Zang et~al.(2026)Zang, Wei, Xu, Wu, Guo, Wang, Lin, Wang, Shi, Xie,
  Xu, Liu, Chen, Tang, Zhang, Zhang, Yu, and Wang]{zang2025rlinfvla}
Hongzhi Zang, Mingjie Wei, Si~Xu, Yongji Wu, Zhen Guo, Yuanqing Wang, Hao Lin,
  Peihong Wang, Liangzhi Shi, Yuqing Xie, Zhexuan Xu, Zhihao Liu, Kang Chen,
  Wenhao Tang, Quanlu Zhang, Weinan Zhang, Chao Yu, and Yu~Wang.
\newblock {RLinf-VLA: A Unified and Efficient Framework for Reinforcement
  Learning of Vision-Language-Action Models}.
\newblock \emph{Robotics: Science and Systems (RSS)}, 2026.
\newblock URL: \url{https://arxiv.org/abs/2510.06710}.

\bibitem[Zhang et~al.(2018)Zhang, Cisse, Dauphin, and
  Lopez-Paz]{zhang2018mixup}
Hongyi Zhang, Moustapha Cisse, Yann~N Dauphin, and David Lopez-Paz.
\newblock {mixup: Beyond Empirical Risk Minimization}.
\newblock \emph{International Conference on Learning Representations (ICLR)},
  2018.
\newblock URL: \url{https://arxiv.org/abs/1710.09412}.

\bibitem[Zhang et~al.(2025{\natexlab{a}})Zhang, Zhang, Jin, Zeng, Li, and
  Wang]{zhang2025robustvla}
Hongyin Zhang, Shuo Zhang, Junxi Jin, Qixin Zeng, Runze Li, and Donglin Wang.
\newblock {RobustVLA: Robustness-Aware Reinforcement Post-Training for
  Vision-Language-Action Models}.
\newblock \emph{arXiv preprint arXiv:2511.01331}, 2025{\natexlab{a}}.
\newblock URL: \url{https://arxiv.org/abs/2511.01331}.

\bibitem[Zhang et~al.(2025{\natexlab{b}})Zhang, Yu, Su, and
  Wang]{zhang2025reinflow}
Tonghe Zhang, Chao Yu, Sichang Su, and Yu~Wang.
\newblock {ReinFlow: Fine-Tuning Flow Matching Policy with Online Reinforcement
  Learning}.
\newblock \emph{Advances in Neural Information Processing Systems (NeurIPS)},
  2025{\natexlab{b}}.
\newblock URL: \url{https://arxiv.org/abs/2505.22094}.

\end{thebibliography}
}

\appendix
\crefalias{section}{appendix}
\crefalias{subsection}{appendix}
\crefalias{subsubsection}{appendix}

\newpage
\section*{\centering Appendix}

\section{Hyperparameters and Experimental Setup}
\label{sec:hyperparams}

This appendix describes what it takes to reproduce the runs behind
\Cref{tab:placement,tab:type,tab:strength}. We provide the post-training and augmentation settings, and details on the evaluation protocol. Every claim in this paper is a comparison between configurations that were
post-trained identically apart from the augmentation, so the settings below are held fixed
across all runs; the only values that vary are the augmentation placement, type and strength
of \Cref{sec:placement,sec:type,sec:strength}.

Code and videos are available on the project page:
\begin{center}
\url{https://bramgrooten.nl/vla-augm/}
\end{center}

\subsection{Hyperparameters}

We post-train with the LIBERO recipes shipped with RLinf~\citep{zang2025rlinfvla}, which
follow the PPO configuration that \citet{chen2025pirl} tuned for flow-matching VLAs, and we
changed as little as possible: an augmentation study is only meaningful if the configurations differ in
the augmentation and nothing else. Where the two models' settings do differ, the difference
was forced by the model rather than chosen by us.

A few choices are worth explaining. We take a \emph{single} update epoch per collected
batch, which is the value RLinf's LIBERO configuration uses; \citet{chen2025pirl}
instead tune this per task, between one and four epochs. Reusing a batch for several epochs
extracts more from each rollout, but it also makes every pass after the first off-policy, so
the clipped importance ratio starts absorbing part of the update. A single epoch keeps every
gradient step on-policy and removes that as a confound, which matters here because the
comparison of interest is between augmentation placements rather than between algorithms.
The learning rates are asymmetric for a related reason: $\pi_{0.5}$ trains its value head at
twenty times the actor's rate, since the head is randomly initialized and has to catch up to
an already competent policy, while GR00T N1.5 uses one rate throughout. Both models keep
their vision--language backbone frozen and train only the action expert together with the
value head, as their supervised fine-tuning recipes do.

The augmentation constants are inherited rather than tuned. The overlay blend $\alpha = 0.5$
and the equal weighting of the clean and augmented critic terms are the values SVEA
uses~\citep{hansen2021svea}, and the $10$-pixel shift is DrQ's $4$-pixel
pad~\citep{yarats2021drq} rescaled to our resolution, as \Cref{sec:framework} describes. We
sweep $\alpha$ in \Cref{sec:strength} precisely because it is the one constant we had no
principled reason to fix. Batch sizes, by contrast, are memory-bound: they are the largest
that fit on the hardware each model ran on. \Cref{tab:hyperparams,tab:aughyper} give the
full settings.

\begin{table}
\centering
\caption{\textbf{VLA post-training hyperparameters.} Settings for the runs of \Cref{sec:experiments}, one column per model. Values shared by both are given once.}
\label{tab:hyperparams}
\small
\setlength{\tabcolsep}{5pt}
\begin{tabular}{lcc}
\toprule
& \emph{$\pi_{0.5}$} & \emph{GR00T N1.5} \\
\midrule
\multicolumn{3}{l}{\emph{Data collection}} \\
Epochs & \multicolumn{2}{c}{$300$} \\
Episodes per epoch & \multicolumn{2}{c}{$512$ ($8$ rounds $\times$ $64$ envs)} \\
Parallel environments & \multicolumn{2}{c}{$64$} \\
Max.\ episode length & \multicolumn{2}{c}{$240$ steps} \\
Observation resolution & \multicolumn{2}{c}{$224 \times 224$, two camera views} \\
Training seed & \multicolumn{2}{c}{$52$} \\
\midrule
\multicolumn{3}{l}{\emph{PPO objective}} \\
Discount $\gamma$ & \multicolumn{2}{c}{$0.99$} \\
GAE $\lambda$ & \multicolumn{2}{c}{$0.95$} \\
Clip ratio (low / high) & \multicolumn{2}{c}{$0.2$ / $0.2$} \\
Dual-clip constant $c$ & \multicolumn{2}{c}{$3.0$} \\
Value clip & \multicolumn{2}{c}{$0.2$} \\
Huber $\delta$ & \multicolumn{2}{c}{$10.0$} \\
Update epochs per batch & \multicolumn{2}{c}{$1$} \\
Advantage normalization & \multicolumn{2}{c}{yes} \\
Entropy bonus / KL penalty & \multicolumn{2}{c}{$0$ / $0$} \\
Reward and log-prob granularity & \multicolumn{2}{c}{action chunk} \\
\midrule
\multicolumn{3}{l}{\emph{Optimization}} \\
Optimizer & \multicolumn{2}{c}{AdamW ($\beta_1 = 0.9$, $\beta_2 = 0.95$)} \\
Actor learning rate & $5 \times 10^{-6}$ & $2 \times 10^{-5}$ \\
Value-head learning rate & $1 \times 10^{-4}$ & $2 \times 10^{-5}$ \\
Adam $\epsilon$ & $1 \times 10^{-8}$ & $1 \times 10^{-5}$ \\
Weight decay & \multicolumn{2}{c}{$0.01$} \\
Gradient clipping & \multicolumn{2}{c}{$1.0$} \\
Global batch size & $2048$ & $1024$ \\
Micro batch size$^\dagger$ & $128$ & $16$ \\
\midrule
\multicolumn{3}{l}{\emph{Model and action head}} \\
Action chunk length & \multicolumn{2}{c}{$5$} \\
Denoising steps & $3$ & $4$ \\
Exploration noise & \multicolumn{2}{c}{flow-SDE~\citep{chen2025pirl}, noise scale $0.5$} \\
Value head input & VLM features & action-head features \\
VLM backbone & \multicolumn{2}{c}{frozen; action expert trained} \\
\midrule
\multicolumn{3}{l}{\emph{Compute}} \\
Hardware (one node) & $4 \times$ H100 & $4 \times$ A100 \\
Sharding & no shard & full shard \\
\bottomrule
\end{tabular}
\vspace{0.4em}

\begin{minipage}{0.72\linewidth}
\footnotesize $^\dagger$Halved to $64$ for $\pi_{0.5}$ in the critic-only and actor-only
runs, which hold a second activation graph. Gradient accumulation absorbs this, so the
global batch size and the optimization are unchanged.
\end{minipage}
\end{table}

\begin{table}[H]
\centering
\caption{\textbf{Augmentation hyperparameters.} Shared by both models. The placement, type
and strength rows are the variables studied in \Cref{sec:placement,sec:type,sec:strength};
their defaults are given here.}
\label{tab:aughyper}
\vspace{0.5em}
\small
\setlength{\tabcolsep}{5pt}
\begin{tabular}{ll}
\toprule
\multicolumn{2}{l}{\emph{Placement and schedule}} \\
Augmented modules & critic only (default) \\
Phases augmented & training updates only; rollouts and evaluations stay clean \\
Augmented views & both cameras (base and wrist) \\
Draw schedule & fresh sample per observation, shared across the two views \\
\midrule
\multicolumn{2}{l}{\emph{Random overlay}} \\
Blend strength $\alpha$ & $0.5$ (default) \\
Critic augmented-term weight $\beta$ & $0.5$ \\
Overlay bank & ImageNet Winter21, $10$ images from each of $19{,}167$ synsets \\
Overlay pre-processing & random resized crop (min.\ area $0.3$) and horizontal flip \\
\midrule
\multicolumn{2}{l}{\emph{Random shift}} \\
Padding & $10$ px ($4.5\%$ of $224$) \\
Padding mode & replicate \\
Per-view sampling & independent shift per camera \\
\bottomrule
\end{tabular}
\end{table}

\subsection{Experimental Setup}

\paragraph{Evaluation.}
The IND column of \Cref{tab:placement,tab:type,tab:strength} is a sweep of the ten LIBERO-Spatial training tasks against $50$ different initial states each, giving $500$ episodes per checkpoint.
The initial states are fixed rather than resampled, so two checkpoints are compared on identical starting conditions. An episode counts as a success if the goal is reached at any point within the horizon (\texttt{success\_once}), the same criterion used
for OOD, so the two columns of a row are directly comparable.

Checkpoints are written every $25$ epochs to test performance regularly. Evaluation is always on clean observations: no augmentation is applied during rollouts or evaluation in any configuration, including the ones whose training updates are augmented.
Action sampling, however, differs between training and evaluation, in the way flow matching allows.
During training the policy explores by taking one of its denoising steps stochastically (\texttt{flow\_sde}~\citep{chen2025pirl}, an analytic noise schedule on top of the flow).
The noise scale of $0.5$ is an amplitude rather than a point in flow time: it multiplies the diffusion coefficient of the SDE, $\sigma(t) = 0.5\sqrt{t / (1 - t)}$, so the standard deviation actually injected at a step is $\sqrt{\Delta t}\,\sigma(t)$.
With three steps that is $0.50$, $0.41$ and $0.20$, decreasing as the chain moves from the prior towards the action, so exploration is strongest where the sample is still mostly noise. At evaluation every step is integrated as a pure ODE
(\texttt{flow\_ode}), which removes that exploration noise and is the
flow-matching counterpart of reporting a deterministic action.
The noise prior drawn, $x_0 \sim \mathcal{N}(0, I)$, remains
a source of randomness even under ODE integration.

\Cref{tab:placement,tab:type,tab:strength} report epoch $300$, the last saved checkpoint, rather than the best epoch per configuration.
The two configurations that collapsed are the exception: they were stopped early and are reported at epoch $100$, which is marked in
\Cref{tab:placement}.

\paragraph{Unweighted averaging.}
The five visual axes differ greatly in size, from $258$ tasks for background textures to
$385$ for objects layout, and the larger axes are also the harder ones. Two ways of
aggregating them therefore disagree. \emph{Micro} pools every episode and takes one mean,
which weights an axis by its task count; \emph{macro} takes each axis's success rate and
averages those five rates unweighted. We report macro average throughout, because the axis sizes
are an artifact of how LIBERO-Plus was assembled rather than a statement about which
shifts matter, and micro would let the two largest axes set the headline number.

\section{Augmentations}
\label{sec:augmentations}

We define our two augmentations and the main settings used in \Cref{sec:framework}. This appendix describes further implementation details behind them and provides more examples.

Augmentation is applied after each model's own preprocessing, at
$224 \times 224$, on observations already normalized to $[-1, 1]$.
Both backbones order their image inputs as (base, left wrist, right wrist), and on LIBERO the third slot is a zero-padded image that the input transform attention-masks away. We therefore augment slots $0$ and $1$ only.

\paragraph{The overlay bank.}
The bank is built once and stored as a single \texttt{uint8} tensor, so training decodes no JPEGs and depends on no dataset filesystem staying mounted.
Images are sampled ten per synset (ImageNet's unit of class: a WordNet synonym
set, one per category, such as ``golden retriever'') rather than uniformly over
files.
This is because ImageNet classes differ greatly in size, and uniform sampling would let the largest ones dominate.
Each image has its short side resized to 224 pixels before a center crop, preserving aspect ratio so that overlays are not distorted.
This yields $184{,}928$ images rather than the $191{,}670$ that ten per synset would give: a synset holding fewer than ten images contributes all it has, and the occasional undecodable file is skipped.

Before each blend, the drawn overlay is itself randomly cropped and flipped: a
rectangle covering at least $30\%$ of \emph{the overlay image's} area is cut out
and resized back up to $224 \times 224$. This crop applies to the stored
ImageNet image only; the observation is never cropped. Because a fresh rectangle
is taken on every draw, the number of distinct overlays a run encounters is far
larger than the number of images stored, and the two are independent knobs: bank
size sets how many \emph{classes} can appear, while the crop sets the variety
\emph{within} one image.

\begin{figure}
\centering
\newlength{\stripw}\setlength{\stripw}{0.93\linewidth}
\setlength{\tabcolsep}{2pt}
\begin{tabular}{@{}c@{\hspace{3pt}}c@{}}
\rotatebox{90}{\footnotesize\ \ \ \ clean} & \includegraphics[width=\stripw]{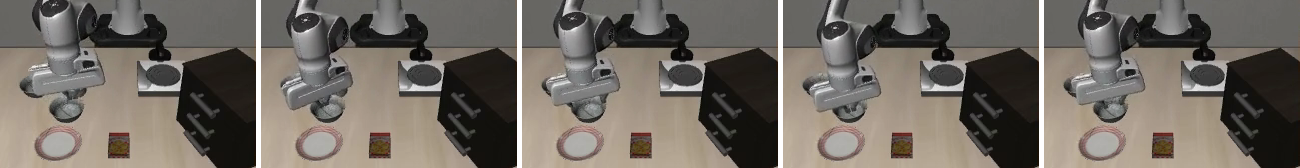} \\[2pt]
\rotatebox{90}{\footnotesize augmented} & \includegraphics[width=\stripw]{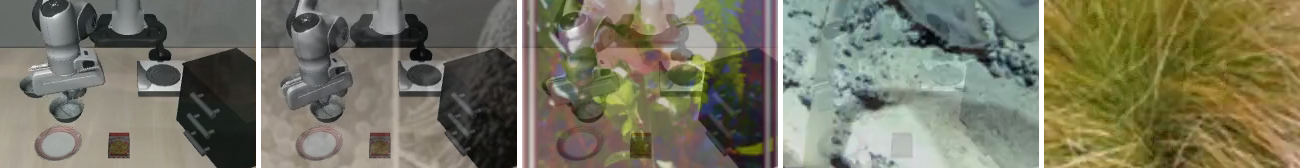} \\[1pt]
& \makebox[\stripw]{%
  \makebox[0.2\stripw]{\footnotesize$\alpha = 0.10$}%
  \makebox[0.2\stripw]{\footnotesize$0.25$}%
  \makebox[0.2\stripw]{\footnotesize$0.50$}%
  \makebox[0.2\stripw]{\footnotesize$0.75$}%
  \makebox[0.2\stripw]{\footnotesize$0.90$}} \\
\end{tabular}
\caption{\textbf{Random overlay across the full strength range.} Five observations from one
LIBERO-Spatial episode, as the policy receives them (top) and after blending with an
ImageNet image (bottom), at the $\alpha$ below each column. Every column draws its own
overlay, randomly resized, cropped and flipped before blending. Our main experiments use
$\alpha = 0.5$, the middle column; \Cref{sec:strength} varies it. Rollouts and evaluations
always use the clean view, while updates (may) receive augmented versions.}
\label{fig:overlay}
\end{figure}

In \Cref{fig:overlay} we show a full sweep across the strength range of the overlay augmentation. Anything above $\alpha = 0.5$ becomes quite difficult, although the critic may still be able to stitch together the current robot pose from slight pixel value deviations.

\paragraph{Placement.}
\Cref{sec:placement} compares four choices of \emph{where} augmented pixels
enter the update. Because $\pi_{0.5}$ and GR00T both run a single backbone pass
whose output feeds the action expert and the value head alike, placement also
decides how many passes an update costs:

\begin{itemize}
\setlength{\itemsep}{1pt}\setlength{\parskip}{0pt}
\item \texttt{none}: no augmentation. One pass; the baseline.
\item \texttt{uniform}: one augmented pass feeding both heads, so actor and
      critic read the same augmented view. Still one pass, which is what makes it
      the naive default.
\item \texttt{critic\_only}: two passes. The actor is
      conditioned on the clean view, and the critic is fit on both views, with
      loss
      $(1 - \beta)\,\mathcal{L}(V_{\text{clean}}) +
      \beta\,\mathcal{L}(V_{\text{aug}})$,
      where $\beta = 0.5$ by default.
\item \texttt{actor\_only}: the mirror image, which isolates \emph{which}
      module the augmentation must reach. Two passes: the actor reads the
      augmented view and the critic a clean one. There is no blending here, since
      the critic sees a single view, so $\beta$ does not apply.
\end{itemize}

\paragraph{Reproducibility.}
Every augmentation draws from seeded generators for sampling images and for the random crops to improve reproducibility.
The two random generators are separate because the image bank may live in host memory while cropping runs on the training device, and a generator may only be used on its own device.

\section{Additional results}
\label{sec:additional_results}

This appendix provides additional insights into the measurements of \Cref{tab:placement,tab:type,tab:strength}.
Everything here is $\pi_{0.5}$ as it is the only model for which we saved checkpoints densely enough (at every $25$ epochs) to plot a curve. Each point is a full sweep of the five visual axes, the same $1662$ tasks the tables score.

\subsection{Performance per OOD category.}
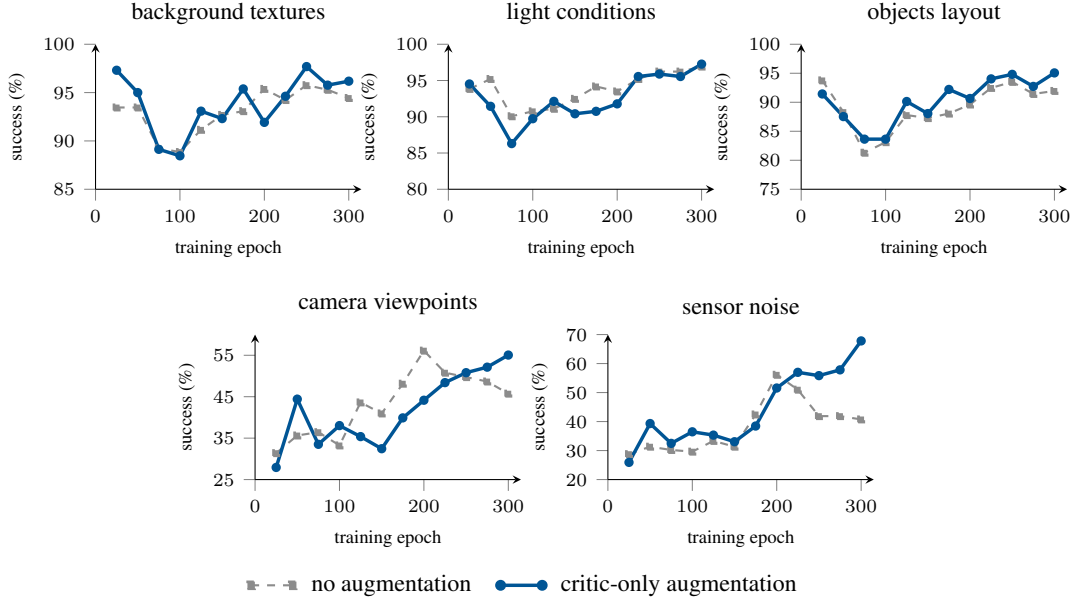
\begin{figure}
\centering
\begin{tikzpicture}
\begin{groupplot}[
  group style={group size=3 by 1, horizontal sep=1.15cm, vertical sep=1.85cm},
  width=0.365\linewidth, height=3.5cm,
  xlabel={training epoch}, ylabel={success (\%)},
  xlabel style={font=\scriptsize}, ylabel style={font=\scriptsize},
  tick label style={font=\scriptsize},
  title style={font=\small, yshift=-2pt},
  xmin=0, xmax=315,
  xtick={0,100,200,300},
  axis lines=left, tick align=outside,
  every axis plot/.append style={mark size=1.1pt},
]
\nextgroupplot[title={background textures}, ymin=85, ymax=100, ytick={85,90,95,100}]
\addplot[color=black!45, mark=square*, dashed, line width=0.8pt] coordinates {(25,93.46) (50,93.46) (75,89.23) (100,88.85) (125,91.15) (150,92.69) (175,93.08) (200,95.38) (225,94.23) (250,95.77) (275,95.29) (300,94.44)};
\addplot[color=ACMDarkBlue, mark=*, solid, line width=1.3pt] coordinates {(25,97.31) (50,95.00) (75,89.13) (100,88.46) (125,93.08) (150,92.31) (175,95.38) (200,91.92) (225,94.62) (250,97.69) (275,95.77) (300,96.18)};
\nextgroupplot[title={light conditions}, ymin=80, ymax=100, ytick={80,85,90,95,100}]
\addplot[color=black!45, mark=square*, dashed, line width=0.8pt] coordinates {(25,93.84) (50,95.21) (75,90.07) (100,90.75) (125,91.10) (150,92.47) (175,94.18) (200,93.49) (225,95.21) (250,96.23) (275,96.23) (300,96.92)};
\addplot[color=ACMDarkBlue, mark=*, solid, line width=1.3pt] coordinates {(25,94.52) (50,91.44) (75,86.30) (100,89.73) (125,92.12) (150,90.41) (175,90.75) (200,91.78) (225,95.55) (250,95.89) (275,95.55) (300,97.26)};
\nextgroupplot[title={objects layout}, ymin=75, ymax=100, ytick={75,80,85,90,95,100}]
\addplot[color=black!45, mark=square*, dashed, line width=0.8pt] coordinates {(25,93.77) (50,88.31) (75,81.30) (100,83.12) (125,87.79) (150,87.27) (175,88.05) (200,89.61) (225,92.47) (250,93.51) (275,91.43) (300,91.95)};
\addplot[color=ACMDarkBlue, mark=*, solid, line width=1.3pt] coordinates {(25,91.43) (50,87.53) (75,83.64) (100,83.64) (125,90.13) (150,88.05) (175,92.21) (200,90.65) (225,94.03) (250,94.81) (275,92.73) (300,95.06)};
\end{groupplot}
\end{tikzpicture}

\vspace{6pt}

\begin{tikzpicture}
\begin{groupplot}[
  group style={group size=2 by 1, horizontal sep=1.15cm, vertical sep=1.85cm},
  width=0.365\linewidth, height=3.5cm,
  xlabel={training epoch}, ylabel={success (\%)},
  xlabel style={font=\scriptsize}, ylabel style={font=\scriptsize},
  tick label style={font=\scriptsize},
  title style={font=\small, yshift=-2pt},
  xmin=0, xmax=315,
  xtick={0,100,200,300},
  axis lines=left, tick align=outside,
  every axis plot/.append style={mark size=1.1pt},
]
\nextgroupplot[title={camera viewpoints}, ymin=25, ymax=60, ytick={25,35,45,55}]
\addplot[color=black!45, mark=square*, dashed, line width=0.8pt] coordinates {(25,31.38) (50,35.64) (75,36.44) (100,33.24) (125,43.62) (150,40.96) (175,48.14) (200,56.12) (225,50.80) (250,49.73) (275,48.67) (300,45.74)};
\addplot[color=ACMDarkBlue, mark=*, solid, line width=1.3pt] coordinates {(25,27.93) (50,44.41) (75,33.51) (100,38.03) (125,35.37) (150,32.45) (175,39.89) (200,44.15) (225,48.40) (250,50.80) (275,52.13) (300,55.05)};
\nextgroupplot[title={sensor noise}, ymin=20, ymax=70, ytick={20,30,40,50,60,70}]
\addplot[color=black!45, mark=square*, dashed, line width=0.8pt] coordinates {(25,28.77) (50,31.34) (75,30.20) (100,29.63) (125,33.33) (150,31.34) (175,42.45) (200,56.13) (225,51.00) (250,41.88) (275,41.88) (300,40.74)};
\addplot[color=ACMDarkBlue, mark=*, solid, line width=1.3pt] coordinates {(25,25.93) (50,39.32) (75,32.48) (100,36.47) (125,35.33) (150,33.05) (175,38.46) (200,51.57) (225,56.98) (250,55.84) (275,57.83) (300,67.81)};
\end{groupplot}
\end{tikzpicture}

\vspace{4pt}

\begin{tikzpicture}[font=\small]
\draw[color=black!45, dashed, line width=0.8pt, mark=square*, mark size=1.4pt] plot coordinates {(0.0,0) (0.55,0)};
\node[anchor=west, inner sep=2pt] at (0.72,0) {no augmentation};
\draw[color=ACMDarkBlue, solid, line width=1.3pt, mark=*, mark size=1.4pt] plot coordinates {(3.3,0) (3.8499999999999996,0)};
\node[anchor=west, inner sep=2pt] at (4.02,0) {critic-only augmentation};
\end{tikzpicture}
\caption{\textbf{Out-of-distribution success during post-training, per perturbation axis.}
The unaugmented $\pi_{0.5}$ PPO baseline against critic-only overlay augmentation at $\alpha = 0.5$, on each visual axis separately.}
\label{fig:axiscurveszoom}
\end{figure}

We present the performance of $\pi_{0.5}$ across each OOD metric in \Cref{fig:axiscurveszoom}. The main OOD average turns out to hide two quite different situations.
On background textures and light conditions the unaugmented baseline is already above $90\%$, so there is almost nothing to win, and the two configurations are barely separable.
On camera viewpoints and sensor noise there is room, and that is where the average gain comes from.
The gain is also late, as we expect the critic needs some additional time to learn through the augmented observations. On both of those axes the unaugmented baseline actually leads through the middle of training, by as much as $12$ points at epoch $200$, and critic-only pulls away only over the final quarter.

\subsection{Training analysis.}
\label{sec:training_analysis}
\begin{figure}
\centering
\input{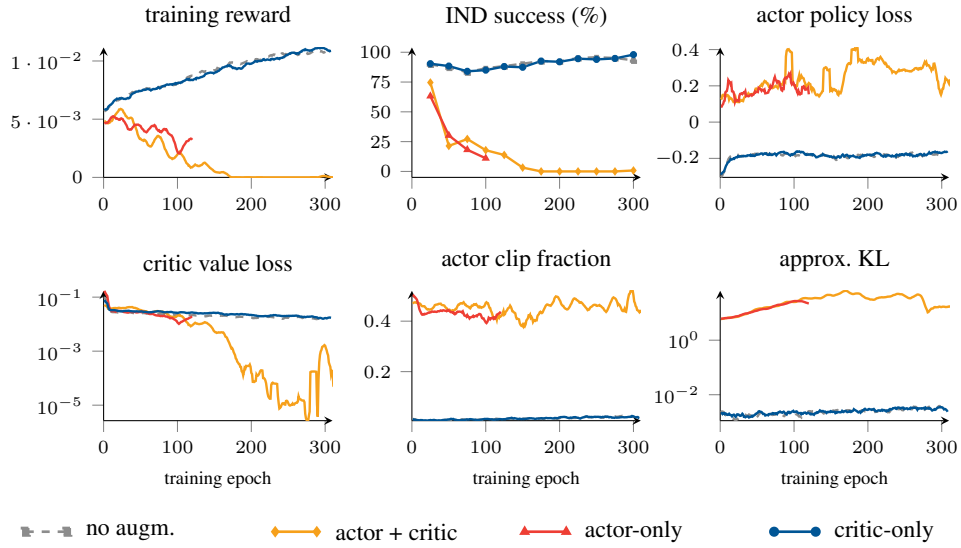}
\caption{\textbf{Training dynamics under each augmentation placement.}
The four $\pi_{0.5}$ placements of \Cref{tab:placement}. Per-epoch traces are smoothed with a centered nine-epoch moving average; evaluations are plotted as measured.}
\label{fig:dynamics}
\end{figure}

We have demonstrated that augmenting the actor leads to training collapse in \Cref{sec:placement}, but not what happened on the way there. \Cref{fig:dynamics} shows the underlying metrics in detail.
Training reward and in-distribution success separate the two groups almost immediately: the actor-augmented runs never go beyond the pretrained VLA performance, while the unaugmented and critic-only runs climb together.
The actor's policy loss runs much higher whenever the actor is augmented, because the augmented observations lead the actor to produce largely varying actions. The approximate KL and clip fraction rise with it, which indicates a policy moving far in each update without the reward following.
Critic-only augmentation raises the value loss slightly above the baseline, as it must when the critic is fit on two views (augmented \textit{and} clean) instead of one, with none of the instability.

\subsection{In-distribution vs OOD success.}
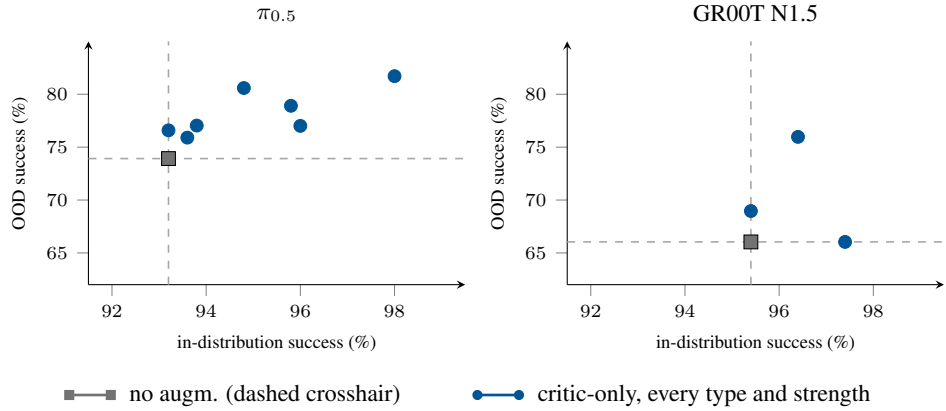
\begin{figure}
\centering
\begin{tikzpicture}
\begin{groupplot}[
  group style={group size=2 by 1, horizontal sep=1.35cm},
  width=0.47\linewidth, height=4.8cm,
  xlabel={in-distribution success (\%)}, ylabel={OOD success (\%)},
  xlabel style={font=\scriptsize}, ylabel style={font=\scriptsize},
  tick label style={font=\scriptsize},
  title style={font=\small, yshift=-2pt},
  xmin=91.5, xmax=99.5, ymin=62, ymax=85,
  xtick={92,94,96,98}, ytick={65,70,75,80},
  axis lines=left, tick align=outside,
  every axis plot/.append style={only marks, mark size=2.4pt},
]
\nextgroupplot[title={$\pi_{0.5}$}]
\draw[black!35, dashed, line width=0.6pt] (axis cs:91.5,73.92) -- (axis cs:99.5,73.92);
\draw[black!35, dashed, line width=0.6pt] (axis cs:93.2,62) -- (axis cs:93.2,85);
\addplot[color=ACMDarkBlue, mark=*, mark options={fill=ACMDarkBlue}] coordinates {(98,81.71) (95.8,78.91) (93.6,75.9) (96,77.01) (93.8,77.04) (94.8,80.59) (93.2,76.59)};
\addplot[color=black, mark=square*, mark size=2.6pt, mark options={fill=black!55}] coordinates {(93.2,73.92)};
\nextgroupplot[title={GR00T N1.5}]
\draw[black!35, dashed, line width=0.6pt] (axis cs:91.5,66.04) -- (axis cs:99.5,66.04);
\draw[black!35, dashed, line width=0.6pt] (axis cs:95.4,62) -- (axis cs:95.4,85);
\addplot[color=ACMDarkBlue, mark=*, mark options={fill=ACMDarkBlue}] coordinates {(97.4,66.04) (96.4,75.97) (95.4,68.96)};
\addplot[color=black, mark=square*, mark size=2.6pt, mark options={fill=black!55}] coordinates {(95.4,66.04)};
\end{groupplot}
\end{tikzpicture}

\vspace{6pt}

\begin{tikzpicture}[font=\small]
\draw[color=black!55, solid, line width=1.0pt, mark=square*, mark size=1.4pt] plot coordinates {(0.0,0.0) (0.55,0.0)};
\node[anchor=west, inner sep=2pt] at (0.72,0.0) {no augm. (dashed crosshair)};
\draw[color=ACMDarkBlue, solid, line width=1.0pt, mark=*, mark size=1.4pt] plot coordinates {(5.4,0.0) (5.95,0.0)};
\node[anchor=west, inner sep=2pt] at (6.12,0.0) {critic-only, every type and strength};
\end{tikzpicture}
\caption{\textbf{In-distribution against out-of-distribution success.}
One marker per configuration at its final checkpoint; the dashed crosshair marks the unaugmented baseline. Note that every critic-only augmentation used maintains or improves both IND and OOD performance.}
\label{fig:tradeoff}
\vspace{2em}
\end{figure}

A natural worry about training against perturbed observations is that robustness is bought with reduced performance on the training distribution.
Plotting the two against each other in \Cref{fig:tradeoff} shows it is not: if that trade were happening, the critic-only points would sit above and to the \emph{left} of the baseline, and instead every one is at or above it with most also to the right.

\end{document}